\documentclass[preprint,12pt,number]{elsarticle}

\usepackage{amssymb}

\usepackage{graphicx}
\usepackage[utf8]{inputenc}
\usepackage{url}
\usepackage[utf8]{inputenc}
\usepackage{booktabs}
\usepackage{multirow} 
\usepackage{float}
\usepackage{booktabs} 
\usepackage[table]{xcolor}
\usepackage{graphicx}
\usepackage{subcaption}
\usepackage{longtable}
\usepackage{tabularx}
\usepackage{array}
\usepackage{ragged2e}
\usepackage{comment}
\usepackage{amsmath} 

\usepackage{rotating}
\journal{}

\begin{document}

\begin{frontmatter}

\title{Behaviour-Conditioned Neural Processes for Adaptive Residential Short-Term Load Forecasting} 

\author[inst1]{Ramin Soleimani\corref{cor1}}
\ead{ramin.soleimani@cs.ucc.ie}

\author[inst1]{Andrea Visentin}
\ead{andrea.visentin@ucc.ie}

\author[inst1]{Dirk Pesch}
\ead{dirk.pesch@ucc.ie}

\cortext[cor1]{Corresponding author}

\affiliation[inst1]{
  organization={School of Computer Science and Information Technology, University College Cork},
  city={Cork},
  country={Ireland}
}

\begin{abstract}

Residential short-term load forecasting (STLF) is challenging because household demand is heterogeneous, temporally variable, and shaped by diverse behavioural routines. Although heterogeneous profiles can support effective global forecasting through statistical sharing, a key modelling challenge is to exploit shared regularities while preserving behaviour-specific variation. This work investigates whether inferred behavioural structure can be embedded within the forecasting mechanism of a Neural Process-based probabilistic model, rather than only as an external grouping signal, for context-conditioned residential STLF across heterogeneous profiles. We propose a behaviour-conditioned Attentive Neural Process framework treating each load profile as a forecasting task. Behavioural structure is represented by a discrete latent variable inferred from the available context and used for behaviour-conditioned decoder conditioning, while a continuous latent variable captures shared functional uncertainty across heterogeneous profiles. To enable conditioning without ground-truth behavioural labels, clustering-derived information provides weak supervision during training, whereas test-time conditioning relies only on context-inferred class distributions. Experiments on the Smart Grid, Smart City (SGSC) dataset use user-disjoint train/validation/test splits, variable context lengths, and multi-step forecast horizons, with comparisons against a label-agnostic Attentive Neural Process (ANP) baseline and fixed-window deterministic STLF baselines. The proposed variants consistently improve MAE and CRPS over ANP across horizons and context settings, with the largest gains under limited context. The best-performing variant achieves average reductions of 7.9\% in MAE and 6.9\% in CRPS relative to ANP. Compared with fixed-window baselines, this variant achieves lower RMSE across all evaluated horizons while maintaining competitive MAE, suggesting fewer large prediction deviations under heterogeneous consumption patterns.These results support single-model, uncertainty-aware forecasting across heterogeneous households, contexts, and horizons.

\end{abstract}

\begin{highlights}
\item We propose a behaviour-conditioned ANP for residential STLF.
\item Context-inferred behavioural structure conditions the forecasting model.
\item Weak clustering supervision enables label-free test-time conditioning.
\item User-disjoint SGSC experiments test variable contexts and horizons.
\item FiLM-ANP-Soft reduces MAE by 7.9\% and CRPS by 6.9\% over ANP.
\end{highlights}


\end{frontmatter}



\section{Introduction}

Modern power systems need accurate short-term load forecasting (STLF) and increasingly benefit from adaptive forecasting capabilities for the reliable and economical operation of applications such as demand response, energy scheduling, and renewable energy integration~\cite{Eren2024}. However, recent developments in smart metering infrastructure, including the widespread use of smart meters and IoT sensors that collect high-frequency consumption data, introduce significant adaptation challenges for STLF, requiring forecasting models that can condition predictions on newly observed data~\cite{kong2019}. Evolving consumer behaviour gives rise to nonlinear and non-stationary load patterns, requiring models with strong representational capacity~\cite{zhang2023_apen}. Together, these developments expose a trade-off in existing STLF methods between representational power and rapid adaptation to new observations. Methods based on simplified load dynamics, such as time-varying linear models derived from ARMA processes, can be updated efficiently in online settings but may struggle with strongly nonlinear consumption patterns~\cite{hasan2025}. Conversely, high-capacity models such as deep neural networks can capture complex usage behaviours but often require retraining or fine-tuning to incorporate new user-specific information, limiting their agility in real-time settings~\cite{gao2023}. This motivates a Neural Process-based forecasting framework that adapts predictions at test time by conditioning on observed load-response pairs while preserving probabilistic uncertainty estimates.

In addition to adaptation, residential STLF must account for the heterogeneity of household electricity demand. Individual load profiles may vary in magnitude, temporal structure, seasonal behavior, and behavioral routines, indicating differences in occupancy, appliance usage, building attributes, and overarching lifestyle consumption patterns~\cite{Lu2015,Quesada2025}. In customer-level load forecasting, such heterogeneity is intricately linked to generalization challenges, encompassing predictions for unseen customers and distribution drift as consumption patterns evolve over time~\cite{Wang2025}.  Heterogeneity does not necessarily preclude effective global forecasting: when related load profiles are modelled jointly, statistical sharing can help a global model learn population-level regularities~\cite{montero2021principles}. However, a single pooled forecasting model must also preserve behaviour-specific variation rather than collapsing distinct consumption regimes into a uniform representation. This creates a central modelling challenge: how to exploit shared structure across households while allowing forecasts to adapt to the behavioural context of each prediction task.

A common approach to structured heterogeneity is to organise load profiles into groups, clusters, typologies, or partitions. Data-driven typologies offer interpretable representations of recurring consumption patterns, while global-local forecasting studies frame partitioning as an intermediate regime between fully local and fully global models~\cite{Quesada2025,montero2021principles}. Recent research in forecasting-oriented clustering indicates that partitions may be evaluated by predictive performance rather than only on geometric similarity~\cite{LopezOriona2025}. Nonetheless, such structure is frequently employed externally, such as for pre-processing labels, fixed partitions, group-specific models, or auxiliary inputs. This motivates incorporating inferred behavioural structure directly into the probabilistic forecasting mechanism of a single global model.

Motivated by this gap, we introduce a behaviour-conditioned Attentive Neural Process framework for residential STLF. The model builds on Neural Processes as probabilistic context-conditioned function learners~\cite{Garnelo2018} and on Attentive Neural Processes, which use attention to improve target-specific conditioning. We formulate forecasting as a context-conditioned function-learning problem, generating predictions from observed context points rather than from a fixed input--output window. A continuous latent variable captures shared functional uncertainty, while a discrete behavioural latent variable represents context-inferred consumption-regime structure. The inferred behavioural variable is mapped to a conditioning representation that modulates a Transformer-style decoder, allowing the predictive distribution to vary with behavioural context within a single global probabilistic model. This modulation-based conditioning follows the general principle of feature-wise conditioning introduced by FiLM~\cite{Perez2018}, with adaptive-normalisation-style conditioning providing a related mechanism for scalable conditional Transformer architectures~\cite{peebles2023}. Clustering-derived information is used only as weak supervision during training, while test-time conditioning relies solely on class distributions inferred from the available context.

\section*{Contributions}
The key contributions of this work are summarised as follows.
\begin{itemize}

    \item \textbf{Behaviour-conditioned Neural Process formulation for residential STLF.}
        We formulate residential short-term load forecasting as a context-conditioned function-learning problem, in which each load profile is treated as a forecasting task conditioned on observed context points. Moving beyond fixed-window input--output mappings and external use of behavioural information as an external grouping signal, the proposed formulation embeds inferred behavioural structure within a unified probabilistic forecasting mechanism.

\item \textbf{Dual latent representation of functional uncertainty and behavioural variation.}
The model separates two sources of variation through complementary latent variables: a continuous latent variable that captures shared functional uncertainty over plausible load trajectories, and a discrete behavioural latent variable that represents context-inferred consumption-regime structure. The behavioural variable conditions the decoder through behaviour-conditioned modulation, while the continuous latent path remains class-agnostic, separating uncertainty modelling from behaviour-aware conditioning within a single probabilistic framework.

\item \textbf{Weakly supervised behavioural inference without test-time labels.}
To enable behaviour-aware conditioning without ground-truth behavioural labels, clustering-derived profile information is used only as weak supervision during training. A label inference network learns to map the available context to a distribution over behavioural classes, so decoder conditioning at test time depends only on observable past data. This prevents future-information leakage from complete-profile cluster labels while retaining behaviour-aware forecasting.

\item \textbf{User-disjoint evaluation across contexts, horizons, and forecasting paradigms.}
We evaluate the proposed framework on the Smart Grid, Smart City (SGSC) residential smart-meter dataset using user-disjoint train/validation/test splits, variable context lengths, and multi-step forecast horizons. The evaluation compares the behaviour-conditioned variants against a label-agnostic ANP baseline and fixed-window deterministic STLF baselines. Results show consistent MAE and CRPS improvements over ANP, with the best-performing variant reducing MAE by 7.9\% and CRPS by 6.9\% on average, while also achieving lower RMSE than the deterministic baselines across all evaluated horizons.

\end{itemize}

The remainder of this paper is organised as follows. Section~2 reviews background concepts, including NPs and related modelling components. Section~3 presents the proposed FiLM-conditioned NPs framework. Section~4 describes the dataset, preprocessing steps, and behavioural clustering approach. Section~5 details the training procedure and experimental setup. Section~6 reports the experimental results and provides a comparative evaluation. Section~7 presents additional analysis and discussion. Finally, Section~8 concludes the paper and discusses future work.


\section{Background and Related Work}

\subsection{Neural Processes (NPs)}
NPs are a class of neural latent variable models that combine features of Gaussian Processes (GPs)~\cite{WilliamsRasmussen1995} and neural networks~\cite{Garnelo2018}. They learn distributions over functions from observed data, which allows them to make probabilistic predictions conditioned on context inputs.Standard GPs usually need $\mathcal{O}(N^3)$ operations because of kernel matrix inversion, but NPs avoid explicit kernel inversion through amortized inference mechanisms~\cite{Kim2019}. As a result, NPs provide a scalable framework for probabilistic function learning by relying on amortized neural inference rather than kernel matrix inversion. NPs were originally used for tasks like regression and black-box optimization.

\subsection{Attentive Neural Processes (ANPs)}
While NPs are efficient and flexible, they can underfit the observed context points, often producing over-smoothed predictions that do not fully explain the data~\cite{Kim2019}. ANPs address this limitation by introducing attention into the NP framework, allowing each target query to attend to the most relevant context observations. This target-dependent conditioning improves predictive accuracy and alleviates the underfitting behaviour of standard NPs~\cite{Kim2019}. As a result, ANPs retain the uncertainty-aware and adaptive nature of NPs while modelling more complex, context-specific patterns. These properties make ANPs well suited to short-term load forecasting, where local temporal dependencies and user-specific consumption patterns are important for accurate and adaptive prediction~\cite{Soleimani2025}.

\subsection{Feature-wise Linear Modulation (FiLM)}
Feature-wise Linear Modulation (FiLM) is a method that adjusts neural network representations using learned feature-wise affine transformations~\cite{Perez2018}. In a FiLM layer, feature-wise scaling and shifting are applied to intermediate activations, with the modulation parameters coming from conditioning information. This approach offers a straightforward way to let additional information, such as labels or context variables, influence how features are processed without changing the network’s structure. FiLM was initially shown to work well for vision-and-language reasoning tasks, such as the CLEVR benchmark~\cite{Perez2018}. Similar modulation-based methods have also been used for sequential data. For instance, Temporal FiLM (TFiLM) uses recurrent conditioning to adjust convolutional representations and capture long-range dependencies in sequences~\cite{Birnbaum2019}. Because of these features, FiLM is a flexible way to add side information to probabilistic and adaptive forecasting models.

\subsection{Adaptive Uncertainty-Aware Modeling}
The above methods aim to produce models that are both adaptive and uncertainty-aware, properties increasingly recognised as essential in modern energy forecasting ~\cite{Eren2024}. NPs naturally support this objective by enabling rapid adaptation to new observations while providing a principled probabilistic representation of predictive uncertainty. FiLM, on the other hand, allows models to condition on exogenous factors by dynamically altering internal computations based on context. In this work, we instantiate this conditioning with an \emph{adaptive layer normalization} variant (AdaLN), which applies FiLM parameters after layer normalization, i.e.,
\[
\mathrm{AdaLN}(h\,|\,c) = (1+\gamma(c))\,\mathrm{LN}(h) + \beta(c),
\]
a design shown to yield stable and scalable conditioning in transformer-style decoders ~\cite{peebles2023}.

\subsection{Related Work in Residential Short-Term Load Forecasting}

Residential short-term load forecasting (STLF) has increasingly moved from traditional statistical models toward machine-learning and deep-learning methods that can learn nonlinear temporal patterns from smart-meter data~\cite{Eren2024}. Recurrent models such as long short-term memory (LSTM) networks have been widely used for household-level STLF and have shown strong performance in single-customer forecasting settings~\cite{kong2019}. To improve learning across volatile residential profiles, pooling-based deep recurrent models train on data from multiple households, exploiting shared structure while reducing overfitting in individual series~\cite{shi2018}. More recent work has considered generalisation across heterogeneous customers and changing data distributions, including approaches based on causal inference, transfer learning, graph neural networks, and spatial-temporal modelling~\cite{Wang2025,lin2021}.

Because residential demand is volatile and difficult to predict, probabilistic STLF is important for decision-making under uncertainty. Probabilistic load forecasting aims to represent predictive distributions rather than only point forecasts, allowing forecast quality to be assessed with proper scoring rules and uncertainty-aware metrics~\cite{hongfan2016}. Existing residential probabilistic forecasting methods include quantile-based approaches, such as boosted additive quantile regression for smart-meter demand and pinball-loss-guided LSTM models for individual loads~\cite{bentaieb2016,wang2019_prob}. Other methods estimate full predictive densities using Gaussian processes, mixture-density neural networks, or Markov-chain mixture distributions~\cite{shepero2018,afrasiabi2020,munkhammar2021}. These methods provide important uncertainty-aware benchmarks, but they are generally formulated as fixed forecasting models rather than context-conditioned function learners that adapt their predictive distributions from variable observed context sets.

A further line of work addresses consumer heterogeneity by grouping customers, daily load shapes, or consumption profiles into representative classes. Clustering has long been used for electricity customer classification and load-pattern grouping, with established methods including K-means, fuzzy clustering, hierarchical clustering, and self-organising maps~\cite{Chicco2006,chicco2012}. In smart-meter studies, behavioural segmentation and consumption typologies have been used to identify recurring patterns in daily routines, peak-demand timing, seasonal behaviour, and broader electricity-use variability~\cite{Kaur2022,Quesada2025}. Several forecasting studies then use these clusters externally, for example by grouping customers with similar behaviour before forecasting, training group-specific models, or combining clustering with deep learning and other predictive models~\cite{Quilumba2015,Fu2018,Han2021,Kim2023,Yu2023}. More recent work also links clustering directly to forecasting performance by constructing partitions according to the accuracy of global forecasting models~\cite{LopezOriona2025}.These studies show that cluster- or category-based structure can help organise heterogeneous load profiles and support forecasting pipelines. However, the resulting behavioural information is typically used as a preprocessing step, fixed partition, group-specific modelling choice, or auxiliary input. In contrast, our work uses inferred behavioural structure as an internal conditioning variable within a probabilistic Neural Process forecasting model.

Taken together, existing residential STLF studies have advanced deep temporal modelling, probabilistic uncertainty estimation, and clustering-based representation of consumer heterogeneity. However, these directions are often treated separately: probabilistic forecasting methods commonly operate as fixed input-output” forecasting models, while clustering-based approaches usually use behavioural structure externally through segmentation, group-specific models, or auxiliary inputs. In contrast, this work formulates residential STLF as context-conditioned probabilistic function learning and uses context-inferred behavioural structure as an internal conditioning variable within a Neural Process forecasting model. This enables behavioural adaptation and uncertainty modelling to be handled within a single global probabilistic framework.

\section{Behaviour-Conditioned Attentive Neural Process Model}

\subsection{Notation Summary}

Table~\ref{tab:notation} summarises the main notation used in the proposed behaviour-conditioned Attentive Neural Process framework. The notation covers the context-target formulation, the deterministic attention path, the continuous latent variable for functional uncertainty, and the discrete behavioural variable used for decoder conditioning.

\label{sec:notation}

\begingroup
\footnotesize
\setlength{\tabcolsep}{4pt}
\renewcommand{\arraystretch}{1.08}

\begin{longtable}{p{2.6cm} >{\RaggedRight\arraybackslash}p{\dimexpr\linewidth-2.6cm-4\tabcolsep\relax}}

\caption{Summary of notation used in the proposed Neural Processes formulation.}

\label{tab:notation}\\
\toprule
\textbf{Symbol} & \textbf{Description} \\
\midrule
\endfirsthead

\toprule
\textbf{Symbol} & \textbf{Description} \\
\midrule
\endhead

\midrule
\multicolumn{2}{r}{\textit{Continued on next page}}\\
\endfoot

\bottomrule
\endlastfoot

\multicolumn{2}{l}{\textit{Data and sets}} \\

$\mathcal D_C$ & Context set $\{(x_i^c,y_i^c)\}_{i=1}^{N_C}$ \\
$\mathcal D_T$ & Target set used for posterior inference during training \\
$X_T$ & Target input locations \\
$Y_T$ & Target outputs \\
$N_C,\,N_T$ & Number of context and target points \\

\midrule
\multicolumn{2}{l}{\textit{Deterministic encoder and attention path}} \\
$h_\theta$ & Deterministic encoder mapping $(x_i^c,y_i^c)\mapsto r_i$ \\
$r_i$ & Context embedding produced by $h_\theta$ \\
$r_C$ & Aggregated context representation \\
$r_t^\star$ & \textbf{Target-specific representation} via cross-attention (ANP) \\
$\alpha_{ti}$ & Attention weight between target $x_t^T$ and context point $x_i^c$ \\
$q(\cdot),\,k(\cdot)$ & Query/key projections used in attention \\
$d_k$ & Dimension of attention keys/queries \\

\midrule
\multicolumn{2}{l}{\textit{Latent path (continuous latent variable)}} \\
$z$ & Continuous latent capturing global functional uncertainty \\
$d_z$ & Dimension of latent variable $z$ \\
$f_{\text{enc}}$ & Latent encoder mapping $[x_i,y_i]\mapsto e_i$ \\
$e_i$ & Latent encoder embedding before aggregation \\
$v_C,\,v_T$ & Set summaries used to parameterize prior/posterior over $z$ \\
$p_\phi(z \mid \mathcal D_C)$ & Context-conditioned prior over $z$ \\
$q_\phi(z \mid \mathcal D_T)$ & Target-conditioned variational posterior over $z$ \\
$\operatorname{diag}(\sigma^2)$ & Diagonal covariance with variances on the diagonal \\

\midrule
\multicolumn{2}{l}{\textit{Label path (categorical latent variable)}} \\
$c$ & Discrete latent behavioural class, $c\in\{1,\dots,K\}$ \\
$K$ & Number of behavioural classes \\
$w_C,\,w_T$ & Set summaries used to parameterize prior/posterior over $c$ \\
$p_\psi(c \mid \mathcal D_C)$ & Context-conditioned categorical prior over $c$ \\
$q_\psi(c \mid \mathcal D_T)$ & Target-conditioned variational posterior over $c$ \\
$\boldsymbol{\pi}_\psi(\cdot)$ & Class probability vector output by label prior network \\

\midrule
\multicolumn{2}{l}{\textit{Decoder, conditioning, and objective}} \\
$\mu_\theta(\cdot)$ & Predictive mean of Gaussian likelihood \\
$\sigma_\theta^2(\cdot)$ & Predictive variance of Gaussian likelihood \\
$\gamma_\ell,\,\beta_\ell$ & FiLM/AdaLN scale and shift for Transformer layer $\ell$ \\
$\boldsymbol{c}$ & One-hot or soft class-probability vector used for decoder conditioning\\
$\mathbf{c}_e$ & Label embedding produced by conditioning MLP \\
$\mathbf{p}$ & Prompt token injected into Transformer decoder \\
$d_{\text{model}}$ & Transformer hidden (model) dimension \\
$\mathrm{KL}(\cdot\|\cdot)$ & Kullback--Leibler divergence \\
$H(\cdot)$ & Entropy \\
$\lambda_{\mathrm{sup}}$ & Weight for supervised cross-entropy term \\
$\beta$ & Weight on the KL term in $\beta$-ELBO objective \\

\end{longtable}
\endgroup
For notational convenience, we use separate parameter symbols for the three main components of the model. We let $\theta$ denote the parameters of the main predictive pathway, including the deterministic encoder, cross-attention mechanism, and decoder/conditional likelihood; $\phi$ denote the parameters of the continuous latent path associated with the stochastic variable $z$; and $\psi$ denote the parameters of the label-inference path associated with the discrete behavioural variable $c$. Thus, we slightly abuse the notation $\theta$ to collectively represent the deterministic and decoding components of the model.

\subsection{Problem Formulation and Neural Processes Foundation}

We formulate short-term load forecasting as a conditional function estimation problem. This approach is consistent with NPs, allowing the model to capture shared structure across users while adapting to context-specific variations. Let $\mathcal D_C=\{(x_i^c,y_i^c)\}_{i=1}^{N_C}$ denote the context set of past observations, where $x_i^c$ is the input feature vector and $y_i^c$ is the observed load. Let $X_T=\{x_t^T\}_{t=1}^{N_T}$ denote the target inputs and $Y_T=\{y_t^T\}_{t=1}^{N_T}$ the corresponding target outputs. The goal is to model the predictive distribution
\begin{equation}
p(Y_T \mid X_T, \mathcal D_C).
\label{eq:goal}
\end{equation}

\paragraph{Context encoding}

A permutation-invariant encoder summarises the context through the deterministic path. Each pair $(x_i^c, y_i^c)$ is mapped by a neural network $h_\theta$ to an embedding $r_i = h_\theta(x_i^c,y_i^c)$, and these embeddings are aggregated by averaging:
\begin{equation}
r_C = \frac{1}{N_C}\sum_{i=1}^{N_C} r_i,
\label{eq:mean-agg}
\end{equation}
which ensures invariance to permutations of $\mathcal D_C$. In the Conditional Neural Processes (CNPs) variant, $r_C$ directly conditions the decoder. In the ANPs variant, target-wise cross-attention is used so that each target query $x_t^T$ attends to the context embeddings $\{r_i\}_{i=1}^{N_C}$, treated as keys and values, to produce a target-specific representation

\[
r_t^\star = \sum_{i=1}^{N_C}\alpha_{ti}\,r_i,
\qquad
\alpha_{ti} = \mathrm{softmax}\!\left(\frac{q(x_t^T)^\top k(x_i^c)}{\sqrt{d_k}}\right)_i
\]

where $q(\cdot)$ and $k(\cdot)$ denote learned query and key projections, respectively, and $d_k$ is the key/query dimension. This allows the model to emphasise the most relevant context points for each target prediction.

\begin{figure*}[t]
\centering
\includegraphics[width=\textwidth]{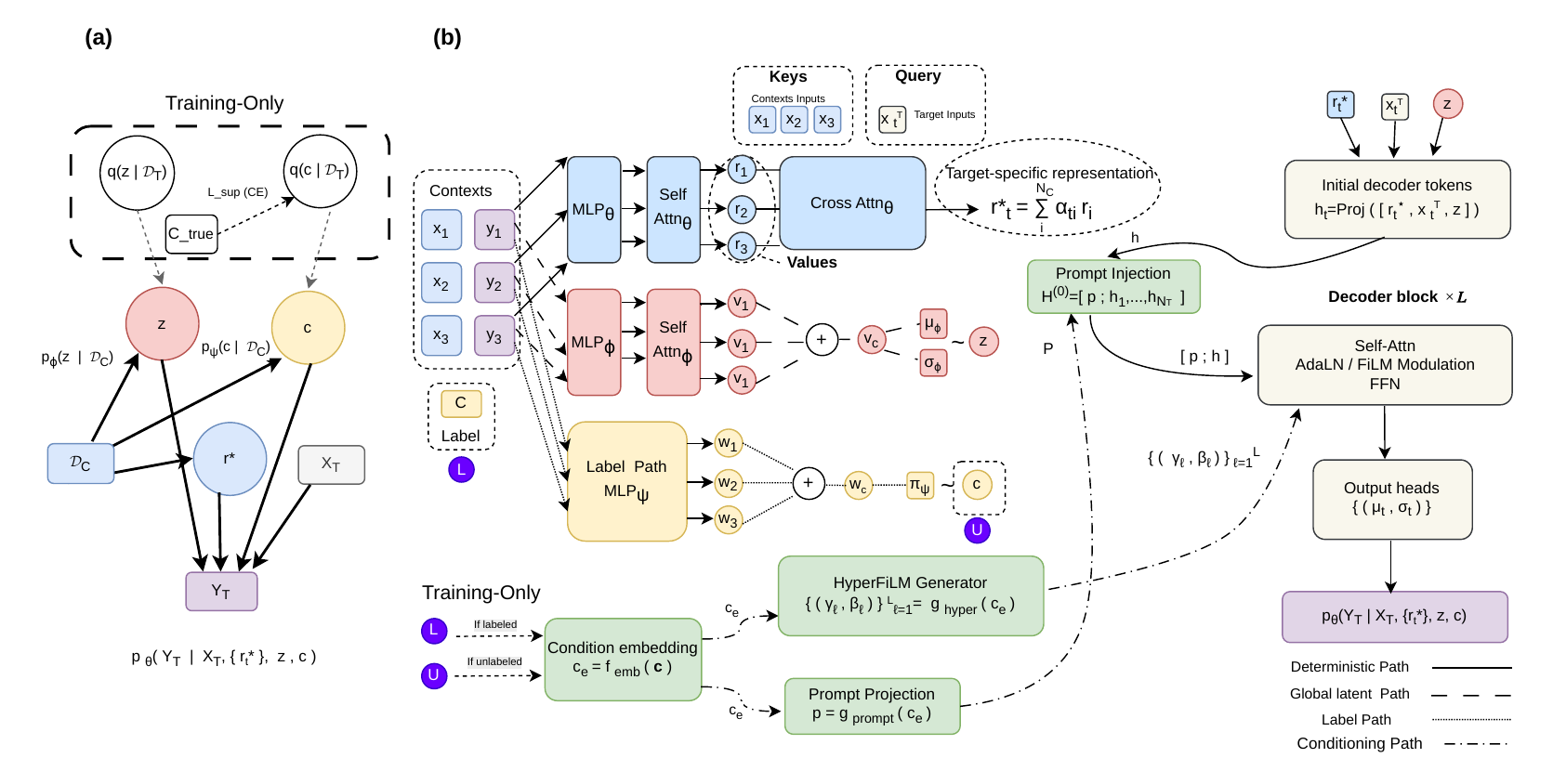}

\caption{Overview of the proposed behaviour-conditioned Attentive Neural Process framework. 
(a) Probabilistic graphical model: a continuous latent variable $z$ captures global functional uncertainty, while a discrete behavioural variable $c$ conditions the decoder without influencing $z$. 
(b) Neural architecture: deterministic attention, latent inference, and behavioural inference feed a conditioned decoder, where the inferred behavioural signal guides decoder conditioning and produces Gaussian predictive outputs.}

\label{fig:model_overview}
\end{figure*}

This deterministic representation corresponds to the upper branch  in Fig.~\ref{fig:model_overview}(b), where the context encoder and cross-attention module produce the target-specific representation $r^\star$ used by the decoder.

\paragraph{Latent path and uncertainty modeling}
To capture global uncertainty over plausible load trajectories, NPs introduce a stochastic latent $z\in\mathbb R^{d_z}$. We use a context-conditioned prior and a target-conditioned posterior,

\begin{equation}
\begin{aligned}
p_\phi(z \mid \mathcal D_C)
&= \mathcal N\!\big(\mu_\phi(v_C),\,\operatorname{diag}\sigma_\phi^2(v_C)\big), \\
q_\phi(z \mid \mathcal D_T)
&= \mathcal N\!\big(\mu_\phi(v_T),\,\operatorname{diag}\sigma_\phi^2(v_T)\big).
\end{aligned}
\label{eq:z-prior-posterior}
\end{equation}

where $v_C$ and $v_T$ are permutation-invariant summaries of $\mathcal D_C$ and $\mathcal D_T$, respectively. These summaries are obtained through a separate latent encoder path. Specifically, a neural encoder $f_{\mathrm{enc}}$, parameterized by $\phi$, maps each concatenated input-output pair to a latent embedding,
\[
e_i = f_{\mathrm{enc}}([x_i,y_i]), \qquad e_i \in \mathbb{R}^{d_e}.
\]

and the set-level summaries are then formed by a permutation-invariant pooling operator,
\[
v_C = \mathrm{Pool}\!\left(\{e_i\}_{(x_i,y_i)\in\mathcal D_C}\right), 
\qquad
v_T = \mathrm{Pool}\!\left(\{e_j\}_{(x_j,y_j)\in\mathcal D_T}\right),
\]
where $v_C, v_T \in \mathbb{R}^{d_v}$ have fixed dimension independent of the number of elements in the set. During training, the posterior set is typically $\mathcal D_T = \mathcal D_C \cup \{(x_t^T,y_t^T)\}_{t=1}^{N_T}$, so $v_T$ summarises both context and target pairs for amortized inference of $q_\phi(z \mid \mathcal D_T)$. At test time, the targets $y$ are unknown, and prediction relies only on the context prior $p_\phi(z \mid \mathcal D_C)$, i.e., on $v_C$ alone. This stochastic latent path corresponds to the global latent branch (red branch) in Fig.~\ref{fig:model_overview}(b).

\paragraph{Decoder and conditional likelihood}
Given $z$ and the context-derived representation, the decoder predicts independent Gaussians:
\begin{equation}
p_\theta(Y_T \mid X_T,\{r_t^\star\}, z)=\prod_{t=1}^{N_T}\mathcal N\!\big(y_t^T;\,\mu_\theta(x_t^T,r_t^\star,z),\,\sigma_\theta^2(x_t^T,r_t^\star,z)\big).
\label{eq:dec-like}
\end{equation}

For scalar targets, $\sigma_\theta^2(\cdot)$ is scalar; for multivariate targets, a diagonal covariance can be used.In Fig.~\ref{fig:model_overview}(b), this operation corresponds to the decoder block on the right-hand side, which combines the deterministic representation, latent sample $z$, and target inputs to produce the predictive distribution.

\paragraph{Training objective}

The generative model for $(Y_T,z)$ given $(X_T,\mathcal D_C)$ is
\begin{equation}
\begin{aligned}
p(Y_T,z \mid X_T,\mathcal D_C)
&= p_\theta\!\bigl(Y_T \mid X_T,\{r_t^\star\},z\bigr)\;
   p_\phi\!\bigl(z \mid \mathcal D_C\bigr).
\end{aligned}
\label{eq:gen-single-z}
\end{equation}

The predictive likelihood is obtained by marginalizing the latent variable $z$ under the context-conditioned prior:
\begin{equation}
p_\theta(Y_T \mid X_T,\mathcal D_C)=\int p_\theta(Y_T \mid X_T,\{r_t^\star\},z)\;p_\phi(z \mid \mathcal D_C)\,dz.
\label{eq:marg-single-z}
\end{equation}

At test time, prediction relies on the context set $\mathcal D_C$ and target inputs $X_T$ only. The deterministic encoder produces the representation $r_j^\star$, the latent variable is sampled from the context-conditioned prior $p_\phi(z\mid\mathcal D_C)$, and the decoder outputs the predictive distribution. During training, the target outputs $Y_T$ are additionally available, allowing the construction of the variational posterior $q_\phi(z\mid\mathcal D_T)$ for amortized inference.

Because the integral in Eq.~\eqref{eq:marg-single-z} is generally intractable, we introduce $q_\phi(z \mid \mathcal D_T)$ and apply Jensen's inequality to obtain the evidence lower bound (ELBO).


\begin{equation}
\begin{aligned}
\log p_\theta(Y_T \mid X_T,\mathcal D_C)
\;\ge\;&
\mathbb E_{q_\phi(z \mid \mathcal D_T)}
\big[\log p_\theta(Y_T \mid X_T,\{r_t^\star\},z)\big] \\
&- \mathrm{KL}\!\big(q_\phi(z \mid \mathcal D_T)\,\|\,p_\phi(z \mid \mathcal D_C)\big)
\end{aligned}
\label{eq:elbo-single-z}
\end{equation}

and the corresponding   $\beta$-weighted training objective
\begin{equation}
\mathcal L_{\text{NP}}
=\mathbb E_{q_\phi(z \mid \mathcal D_T)}\big[\log p_\theta(Y_T \mid X_T,\{r_t^\star\},z)\big]
-\beta\,\mathrm{KL}\!\big(q_\phi(z \mid \mathcal D_T)\,\|\,p_\phi(z \mid \mathcal D_C)\big).
\label{eq:beta-elbo-single-z}
\end{equation}

While the above subsection reviews the standard NP formulation, the proposed approach extends it with an additional discrete behavioural latent variable and a decoder conditioning mechanism. Fig~\ref{fig:model_overview} summarises the resulting probabilistic formulation and its neural implementation.


\subsection{FiLM-Conditioned Neural Processes with Dual Latent Variables}

\paragraph{Overview}

Fig.~\ref{fig:model_overview} summarises the probabilistic formulation and neural architecture of the proposed model. The model extends the standard NP by introducing two latent variables: a continuous latent variable $z$ that captures global functional uncertainty and a discrete latent variable $c \in \{1,\dots,K\}$ that represents behavioural regimes across load profiles. The continuous latent variable $z$ remains class-agnostic and captures residual functional uncertainty, so $c$ and $z$ are modeled independently in both the prior and posterior. The inferred behavioural variable modulates the decoder through FiLM-based conditioning, enabling the model to adapt its feature processing according to the behavioural context inferred from observed history.

The design is flexible with respect to label availability at both training and test time. A KL alignment term and a Gumbel-Softmax relaxation during training allow missing labels to be handled via a semi-supervised ELBO with a categorical KL alignment term for $c$~\cite{jang2017gumbel,maddison2017concrete}. This enables differentiable sampling of the discrete label variable during optimization, augmented with cross-entropy on labeled profiles. During testing, when $c$ is unavailable, the label path provides an estimate so the decoder remains label-conditioned without altering the NP formulation.

\paragraph{Generative model (context-conditioned priors)}

Let $c\in\{1,\dots,K\}$ denote a discrete latent variable representing a behavioural mode. The label path produces a context summary $w_C$ and defines the context-conditioned prior
\[
p_\psi(c\mid\mathcal D_C)=\mathrm{Cat}\!\big(c;\,\boldsymbol\pi_\psi(w_C)\big).
\]
The variable $c$ is used only for decoder conditioning and does not influence the continuous latent variable $z$. The joint distribution over $(Y_T,z,c)$, conditioned on $X_T$ and $\mathcal D_C$, factorizes as
\begin{equation}
p(Y_T,z,c \mid X_T,\mathcal D_C)=p_\theta(Y_T \mid X_T,\{r_t^\star\},z,c)\;p_\phi(z \mid \mathcal D_C)\;p_\psi(c \mid \mathcal D_C),
\label{eq:gen-dual}
\end{equation}
with likelihood
\begin{equation}
p_\theta(Y_T \mid X_T,\{r_t^\star\},z,c)=\prod_{t=1}^{N_T}\mathcal N\!\big(y_t^T;\,\mu_\theta(x_t^T,r_t^\star,z,c),\,\sigma_\theta^2(x_t^T,r_t^\star,z,c)\big).
\label{eq:like-dual}
\end{equation}
The priors $p_\phi(z \mid \mathcal D_C)$ and $p_\psi(c \mid \mathcal D_C)$ are assumed independent.

We use factorized target-conditioned posteriors
\[
q(z,c\mid \mathcal D_T)
=
q_\phi(z\mid \mathcal D_T)\,
q_\psi(c\mid \mathcal D_T),
\]
where
\[
q_\phi(z\mid \mathcal D_T)
=
\mathcal N\!\left(\mu_\phi(v_T),
\operatorname{diag}(\sigma_\phi^2(v_T))\right),
\qquad
q_\psi(c\mid \mathcal D_T)
=
\operatorname{Cat}\!\left(c;\pi_\psi(w_T)\right).
\]
Here, \(w_T\) is the target-set label summary and
\[
\pi_\psi(w_T)=\operatorname{softmax}(a_\psi(w_T))
\]
is the \(K\)-dimensional posterior class-probability vector. During training, these posteriors are used for amortized inference. When a differentiable sample of the class variable is required, we draw a relaxed class vector
\[
\tilde{\mathbf c}
=
\operatorname{GumbelSoftmax}\!\left(a_\psi(w_T),\tau\right)
\in \Delta^{K-1},
\]
where \(\tau\) is the temperature. This relaxed vector is used as the differentiable representation of the categorical variable in decoder conditioning and Monte Carlo estimates of the expectation~\cite{jang2017gumbel,maddison2017concrete}. At test time, the target outputs are unavailable, so prediction relies on the context-conditioned priors \(p_\phi(z\mid\mathcal D_C)\) and \(p_\psi(c\mid\mathcal D_C)\); the label prior may be used either through its soft probability vector or through the corresponding argmax class representation.

\paragraph{Labeled vs.\ unlabeled single-profile objectives}

We first derive the labelled and unlabelled objectives for a single load profile. For clarity, when these objectives are later summed over the training set, $\mathcal L_{\mathrm{lab}}^{(j)}$ and $\mathcal L_{\mathrm{unlab}}^{(j)}$ denote the corresponding objectives evaluated for profile $j$.  If a profile’s label $c$ is known (labeled), there is no variational distribution over $c$. We maximize the joint likelihood $\log p(Y_T,c\mid X_T,\mathcal D_C)$ so that, in addition to the data term, the context-conditioned label prior $p_\psi(c\mid\mathcal D_C)$ is directly trained via $\log p_\psi(c\mid\mathcal D_C)$. The conditional marginal is

\begin{equation}
p(Y_T,c\mid X_T,\mathcal D_C)=p_\psi(c \mid \mathcal D_C)\!\int p_\theta(Y_T \mid X_T,\{r_t^\star\},z,c)\;p_\phi(z \mid \mathcal D_C)\,dz,
\label{eq:marg-labeled}
\end{equation}

and applying Jensen’s inequality with $q_\phi(z \mid \mathcal D_T)$ yields the labeled ELBO:

\begin{align}
\log p(Y_T, c \mid X_T, \mathcal D_C)
\;\ge\;&
\mathcal L_{\mathrm{lab}}^{(j)} \notag\\
={}&
\mathbb E_{q_\phi(z \mid \mathcal D_T)}
\!\left[\log p_\theta\!\left(Y_T \mid X_T, \{r_t^\star\}, z, c\right)\right] \notag\\
&- \mathrm{KL}\!\left(q_\phi(z \mid \mathcal D_T)\,\|\,p_\phi(z \mid \mathcal D_C)\right) \notag\\
&+ \log p_\psi(c \mid \mathcal D_C).
\label{eq:elbo-labeled}
\end{align}

More explicitly, because $c$ is observed, we introduce only the auxiliary posterior $q_\phi(z \mid \mathcal D_T)$ and apply Jensen’s inequality to obtain a variational lower bound as commonly done in variational inference ~\cite{bishop2006prml,kingma2014autoencoding}.

\[
\begin{aligned}
\log p(Y_T,c \mid X_T,\mathcal D_C)
={}&\log p_\psi(c \mid \mathcal D_C) \\
&+ \log \int
p_\theta(Y_T \mid X_T,\{r_t^\star\},z,c)\,
p_\phi(z \mid \mathcal D_C)\,dz .
\end{aligned}
\]

Here, no $c$–KL appears; the single KL term regularizes the target-conditioned $q_\phi(z \mid \mathcal D_T)$ toward the learned amortized context prior $p_\phi(z \mid \mathcal D_C)$, and the term $\log p_\psi(c \mid \mathcal D_C)$ simply rewards the model when the context predicts that the observed class $c$ was likely. This is the standard labeled bound from the semi-supervised VAE that we adapted to the NP setting by conditioning the prior on $\mathcal D_C$,cf.~\cite[Eq.~(6)]{kingma2014semi}
.The reader interested in a detailed derivation of the loss function for amortized variational latent-function models is referred to ~\cite{zhang2019_vi}.

\paragraph{Unlabeled case (intuition and derivation)}
When the label is unknown, we must account for \emph{all} classes and \emph{all} latent $z$ values suggested by the context–conditioned priors. We therefore start from the exact marginal and optimize a variational lower bound of this quantity:
\begin{equation}
p(Y_T \mid X_T,\mathcal D_C)=\sum_{c=1}^K\int p_\theta(Y_T \mid X_T,\{r_t^\star\},z,c)\;p_\phi(z \mid \mathcal D_C)\;p_\psi(c \mid \mathcal D_C)\,dz.
\label{eq:marg-unlabeled}
\end{equation}

This averages the class–conditioned likelihood over $p_\psi(c\mid\mathcal D_C)$ 
and integrates latent uncertainty under $p_\phi(z\mid\mathcal D_C)$.

\paragraph{Variational lower bound}

We introduce a mean–field posterior $q_\phi(z \mid \mathcal D_T)\,q_\psi(c \mid \mathcal D_T)$, rewrite the marginal as an expectation under this posterior, and apply Jensen’s inequality (cf.\ M2,~\cite{kingma2014semi}, Eqs.\ (6)--(7)) to obtain the unlabeled ELBO:

\begin{align}
\log p(Y_T \mid X_T,\mathcal D_C)
\;\ge\;
\mathcal L_{\mathrm{unlab}}^{(j)}
&:=
\mathbb{E}_{q_\phi(z \mid \mathcal D_T)\,q_\psi(c \mid \mathcal D_T)}
\!\Bigl[
\log p_\theta\!\bigl(Y_T \mid X_T,\{r_t^\star\},z,c\bigr)
\Bigr]
\notag\\
&\quad
- \mathrm{KL}\!\left(q_\phi(z \mid \mathcal D_T)\,\|\,p_\phi(z \mid \mathcal D_C)\right)
\notag\\
&\quad
- \mathrm{KL}\!\left(q_\psi(c \mid \mathcal D_T)\,\|\,p_\psi(c \mid \mathcal D_C)\right).
\label{eq:elbo-unlabeled}
\end{align}
\emph{Term-wise interpretation:} (i) the data term averages the log-likelihood over the inferred classes $q_\psi(c\mid\mathcal D_T)$ and the latent $q_\phi(z\mid\mathcal D_T)$; (ii) the $z$-KL pulls the \emph{class-agnostic} posterior $q_\phi(z\mid\mathcal D_T)$ toward the context prior $p_\phi(z\mid\mathcal D_C)$; (iii) the $c$-KL aligns $q_\psi(c\mid\mathcal D_T)$ with the context prior $p_\psi(c\mid\mathcal D_C)$. Because $z$ is class-agnostic in our model (neither prior nor posterior depends on $c$), the $z$-KL appears once rather than inside an expectation over $c$.

\paragraph{Entropy form}
Using $-\mathrm{KL}(q\|p)=\mathbb{E}_{q}[\log p]+H(q)$ for categoricals, the $c$-KL in \eqref{eq:elbo-unlabeled} can be rewritten to make the trade-off explicit (M2,~\cite{kingma2014semi}, Eq.~(7)):

\begin{equation}
\begin{aligned}
\mathcal{L}_{\mathrm{unlab}}^{(j)}
&=
\mathbb{E}_{q_\psi(c \mid \mathcal{D}_T)\,q_\phi(z \mid \mathcal{D}_T)}
\Bigl[
\log p_\theta\bigl(Y_T \mid X_T,\{r_t^\star\},z,c\bigr)
\Bigr]
\\
&\quad
- \mathrm{KL}\!\left(
q_\phi(z \mid \mathcal{D}_T)
\,\|\, 
p_\phi(z \mid \mathcal{D}_C)
\right)
\\
&\quad
+ \mathbb{E}_{q_\psi(c \mid \mathcal{D}_T)}
\Bigl[
\log p_\psi\bigl(c \mid \mathcal{D}_C\bigr)
\Bigr]
\\
&\quad
+ H\!\bigl(q_\psi(c \mid \mathcal{D}_T)\bigr).
\end{aligned}
\label{eq:elbo-unlabeled-entropy}
\end{equation}

where \(H(q_\psi(c\mid\mathcal D_T))\) denotes the entropy of the categorical posterior over \(c\), defined as
\[
H(q_\psi(c\mid\mathcal D_T))
=
-\sum_{k=1}^{K}
q_\psi(c=k\mid\mathcal D_T)
\log q_\psi(c=k\mid\mathcal D_T).
\]

In words, the \(c\)-term rewards agreement with the context-implied class prior while preserving uncertainty through the entropy term when the targets do not strongly support a single class.

\paragraph{Key differences}
When $c$ is known, it is used as an observed conditioning variable and no variational distribution over $c$ is introduced; hence, the labelled objective contains no $c$-KL or entropy term. When $c$ is unknown, the objective follows the semi-supervised variational treatment of Kingma et al.~\cite{kingma2014semi}, in which the data term is averaged over the inferred categorical posterior $q_\psi(c\mid\mathcal D_T)$ and an additional categorical regularizer
\[
-\mathrm{KL}\!\left(q_\psi(c\mid\mathcal D_T)\,\|\,p_\psi(c\mid\mathcal D_C)\right)
\]
aligns the target-conditioned posterior with the context-conditioned prior. Equivalently, this term can be written as an expected log-prior plus the entropy of $q_\psi(c\mid\mathcal D_T)$, which preserves uncertainty when the targets do not strongly support a single class.

\paragraph{Supervised augmentation (labeled profiles)}
We may add a cross-entropy term on labeled data to train the classifier head,
\begin{equation}
\mathcal L_{\mathrm{sup}}^{(j)}=-\log q_\psi\!\big(c_{\mathrm{true}}\mid \mathcal D_T\big),
\label{eq:sup-term}
\end{equation}

where $c_{\mathrm{true}}$ denotes the observed class label. The augmented labelled objective is then
\begin{equation}
\tilde{\mathcal L}_{\mathrm{lab}}^{(j)}
=
\mathcal L_{\mathrm{lab}}^{(j)}
-
\lambda_{\mathrm{sup}}\,\mathcal L_{\mathrm{sup}}^{(j)},
\label{eq:lab-plus-sup}
\end{equation}
where $\lambda_{\mathrm{sup}}\ge 0$ is a weighting coefficient controlling the strength of the supervised classification term. In implementation, this maximization is performed by minimizing the negative objective.

\paragraph{Overall training objective and optimization}

Let $\mathcal J_L$ and $\mathcal J_U$ denote the labeled and unlabeled profile index sets, respectively. Using the profile-specific objectives defined above, the total objective over the training set is

\begin{equation}
\mathcal L_{\mathrm{total}}(\theta,\phi,\psi)
=
\sum_{j\in \mathcal J_L}
\tilde{\mathcal L}_{\mathrm{lab}}^{(j)}
+
\sum_{j\in \mathcal J_U}
\mathcal L_{\mathrm{unlab}}^{(j)}.
\label{eq:total-loss}
\end{equation}
We optimize with stochastic gradients, reparameterizing $z$ and using a differentiable relaxation (e.g., Gumbel--Softmax) for $c$.  

\subsection{Conditional Transformer Decoder with HyperFiLM and Prompt Injection}

The conditional Transformer decoder uses two complementary conditioning mechanisms: HyperFiLM modulation and prompt token injection. Given a one-hot or soft label vector $\boldsymbol{c}\in\mathbb{R}^{K}$, we first compute
\[
\boldsymbol{c}_e = f_{\mathrm{emb}}(\boldsymbol{c}) \in \mathbb{R}^{d_c}.
\]

A compact hypernetwork maps this embedding to layer-specific FiLM parameters $(\gamma_\ell,\beta_\ell)$ for each Transformer layer $\ell$. These parameters modulate normalized activations in the self-attention and feedforward sublayers via

\[
\tilde{h}^{(\ell)} = (1 + \gamma_\ell)\odot \mathrm{LN}(h^{(\ell)}) + \beta_\ell,
\]

where $\odot$ denotes element-wise multiplication. The same conditioning embedding is also projected into the model space to form a single prompt token, which is prepended once to the initial decoder sequence before self-attention, allowing label information to influence the target sequence through attention. These two mechanisms provide  conditioning at the activation and sequence levels.

This embedding is shared across two conditioning branches, as illustrated in Fig.~\ref{fig:model_overview}(b): (i) a HyperFiLM branch that generates layer-wise FiLM parameters via the FiLM generator, and (ii) a prompt-injection branch that feeds a conditioning token directly into the Transformer decoder.

\paragraph{1. HyperFiLM for Per-Layer Modulation}

A hypernetwork $g_{\text{hyper}}$ maps the label embedding $\mathbf{c}_e$ to a stack of FiLM parameters for each of the $L$ Transformer layers:
\[
\left\{ \gamma_\ell, \beta_\ell \right\}_{\ell=1}^L = g_{\text{hyper}}(\mathbf{c}_e), \quad \gamma_\ell, \beta_\ell \in \mathbb{R}^{d_{\text{model}}}
\]

These parameters modulate the output of LayerNorm within each attention and feedforward sublayer (FFN) as follows:
\[
\tilde{\mathbf{h}}^{(\ell)} = (1 + \gamma_\ell) \odot \mathrm{LN}(\mathbf{h}^{(\ell)}) + \beta_\ell
\]

This design provides layer-wise label-dependent conditioning throughout the decoder depth.

\paragraph{2. Prompt Token Injection}

The same embedding $\mathbf{c}_e$ is projected into the model space via a learned linear transformation:

\[
p = g_{\mathrm{prompt}}(c_e) \in \mathbb{R}^{d_{\mathrm{model}}}.
\]

This projected vector is treated as a learnable prompt token and prepended once to the initial decoder sequence before the first self-attention layer. The resulting decoder input is

\[
H^{(0)}=[p;h_1,h_2,\ldots,h_{N_T}]
\in\mathbb{R}^{(N_T+1)\times d_{\mathrm{model}}}.
\]
The prompt token allows label-conditioned information to influence all positions through self-attention~\cite{Lester2021}. The prompt token participates in the decoder stack as part of the sequence representation and is discarded before the output heads, so that only the target positions contribute to the predictive distribution.

\section{Dataset and Preprocessing}
\label{sec:dataset}

We use high-resolution smart meter data from the Smart Grid, Smart City (SGSC) trial in Australia~\cite{SGSC2014}, which provides half-hourly electricity consumption records (in kilowatt-hours, kWh) from over 10{,}000 residential households. For this study, we curate a subset by randomly selecting 400 residential users whose electricity usage in 2013 is fully complete and free from missing values or time gaps. To ensure reproducibility, the selection is performed using a fixed random seed. The number of selected users reflects a practical trade-off between capturing sufficient behavioural diversity and maintaining a consistent evaluation setting across all models. This supports a fair comparison between the proposed global model and conventional forecasting approaches under a unified experimental protocol.

The raw meter readings are processed into a clean consumption matrix of shape \(T \times N\), where \(T = 17{,}520\) denotes the number of 30-minute intervals in a non-leap year and \(N = 400\) is the number of selected users. Each column contains the continuous kWh consumption for one user, with exact time alignment across all users to enable synchronized sequence modeling.

To capture both daily and intra-day load patterns, we segment each user’s time series into overlapping profiles of 104 half-hour steps, comprising two full consecutive days (96 steps) followed by an additional 4-hour segment (8 steps) from the subsequent day.
 
This design
allows short-horizon forecasts to be evaluated at the end of each frame using
complete target segments, resulting in a consistent and well-defined evaluation
protocol across all horizons. These profiles are generated using a sliding window with a stride of one day. Each user’s annual time series is first normalised (z-score) , and segmented into fixed-length windows. Specifically, we extract windows of  length \$W = 104\$ steps with a stride of 48 time steps.
Importantly, while the sliding window defines how training samples are constructed, it does not restrict the prediction start time. Within each window, the forecasting task is defined by randomly sampling a context set and a target segment, such that predictions can be made from arbitrary points within the window depending on the selected context length. This function-space formulation ensures that the model learns to predict from variable starting points rather than from a fixed temporal alignment.

\begin{table}[H]
\centering
\caption{K-means clustering summary on  load profiles as a function of the number of clusters $K$. The final selection is based on the composite rank score, with the lowest rank sum indicating the selected number of clusters.}
\label{tab:kmeans-k-selection-104}
\begin{tabular}{cccccc}
\toprule
$K$ & Inertia & Silhouette & Davies-Bouldin & Calinski-Harabasz & Rank sum \\
\midrule
2 & 4{,}772{,}921.5 & \textbf{0.0560} & 4.1728 & \textbf{2803.83} & 14 \\
\rowcolor{gray!10}
\textbf{3} & 4{,}595{,}137.0 & 0.0424 & 3.7568 & 2423.32 & \textbf{13} \\
4 & 4{,}478{,}501.0 & 0.0402 & 3.8376 & 2091.60 & 15 \\
5 & 4{,}385{,}823.5 & 0.0382 & 3.6405 & 1865.94 & 14 \\
6 & 4{,}307{,}707.0 & 0.0357 & 3.5663 & 1701.10 & 14 \\
7 & 4{,}245{,}248.0 & 0.0344 & \textbf{3.5451} & 1561.00 & 14 \\
\bottomrule
\end{tabular}
\end{table}


\begin{figure*}[t]
    \centering
    \includegraphics[width=\columnwidth]{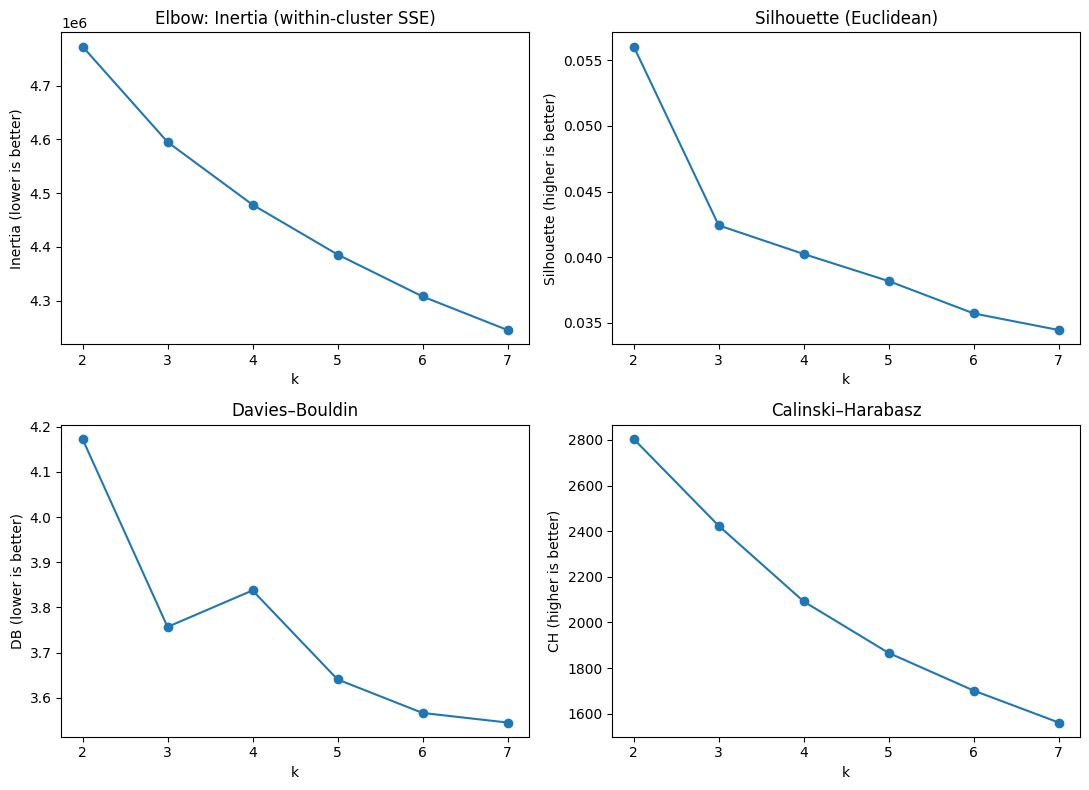}
    \caption{K-means $K$-selection criteria (inertia, silhouette, Davies--Bouldin,
and Calinski--Harabasz) evaluated on  load profiles.}
    \label{fig:kmeans-k-selection-104}
\end{figure*}

\paragraph{Soft Probabilistic Label Encoding}

\begin{figure}[t]
    \centering
    \includegraphics[width=\columnwidth]{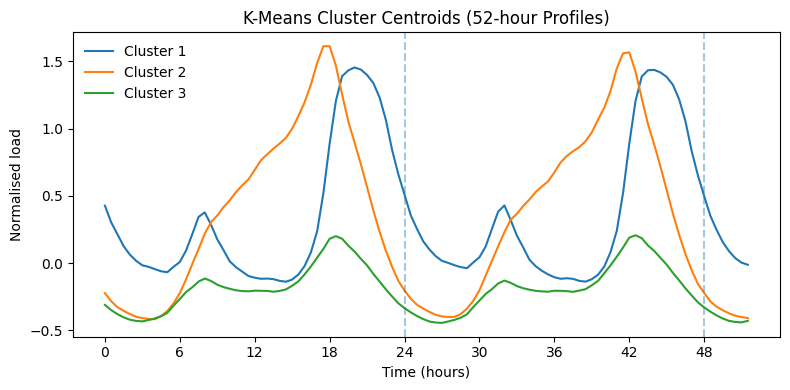}
    \caption{Representative normalised load profiles given by K-means cluster
centroids (52-hour window).}
    \label{fig:kmeans-centroids}
\end{figure}

The K-means clustering step yields three centroids, which we interpret as behavioural archetypes. Fig.~\ref{fig:kmeans-centroids} shows the average profile of each cluster over the 104 half-hour steps. These centroids highlight the main behavioural patterns in the data, but many individual profiles fall between clusters or share features from multiple archetypes. To reflect this, we turn the K-means results into a soft probabilistic label for each frame.

For a profile \(x\), we first compute its Euclidean distance to each centroid,
\[
d_k(x) \;=\; \big\|x - \mu_k\big\|_2, \qquad k=1,\dots,K,
\]
\[
\tilde d_k(x) \;=\; \frac{d_k(x)}{\sqrt{L}},
\]
and then map these normalised distances to a probability vector using a temperature-scaled softmax,
\begin{equation}
p_k(x)
\;=\;
\frac{\exp\!\big(-\tilde d_k(x) / \tau\big)}{\sum_{j=1}^K \exp\!\big(-\tilde d_j(x) / \tau\big)},
\qquad k=1,\dots,K,
\label{eq:soft-kmeans-label}
\end{equation}

where \(\tau\) controls the sharpness of the assignments.

In our experiments, we set $\tau = 0.15$. This value produces soft assignments that are informative but not degenerate. We assess the quality of the resulting soft labels through two complementary properties: consistency with the hard K-means assignment and controlled uncertainty. In every case, the hard cluster index coincides with \(\arg\max_k p_k(x)\), ensuring that the probabilistic encoding remains fully faithful to the original clustering. At the same time, the mean entropy of the soft labels is \(0.7107\) for \(K=3\), which lies well below the maximum value \(\log 3 \approx 1.10\). This shows that the assignments are neither degenerate nor overly diffuse. Each profile has a clear main archetype, but there is still enough uncertainty to represent mixed or transitional behaviours.

The resulting vector $p(x) = (p_1(x),\dots,p_K(x))$ lies on the probability simplex i.e.,  $p_k(x) \geq 0$ for all $k$ and $\sum_{k=1}^{K} p_k(x) = 1$, and encodes the degree of membership of profile $x$ to each behavioural centroid.Well-separated profiles yield peaked distributions, whereas ambiguous or transitional profiles produce more diffuse assignments. This soft representation is beneficial for the proposed conditional architecture for two reasons. First, it can represent mixed behavioural patterns that would be forced into an arbitrary boundary under hard labels. Second, it provides a smooth conditioning signal for FiLM modulation, so small changes in the input profile induce correspondingly small changes in the decoder conditioning.

\section{Model Training Setup}

\subsection{Function-Space Training Regime}

We train the proposed FiLM-conditioned Attentive Neural Processes (FiLM-ANP) in the NP function-space regime. Rather than learning a fixed input--output mapping, the model learns a conditional distribution over functions from partial observations. Each two-day load profile is treated as a discrete realization of an underlying function
\[
f:\{0,\dots,103\}\rightarrow\mathbb{R},
\]
representing a single household consumption trajectory over a 52-hour window with \(L=104\) half-hourly points. Here, \(x_t\) denotes the normalized time coordinate and \(y_t\) the \(z\)-normalized electricity demand at step \(t\). In the NP formulation, both context and target points are sampled from the same underlying load curve, but only a subset is revealed to the model as context.

For each training episode, we sample a context length \(C \sim \mathrm{Unif}\{48,\dots,96\}\). Given a profile length \(L=104\), this leaves \(L-C\) future points available, from which we sample a target horizon \(M \sim \mathrm{Unif}\{1,\dots,L-C\}\). This ensures that the model always observes at least one full day of historical consumption (48 half-hour steps), while exposing it to a variable number of future points for prediction. Since \(L=104\), the maximum available horizon ranges from 56 steps when \(C=48\) to 8 steps when \(C=96\).

The first \(C\) points of each profile, \(\{(x_t,y_t)\}_{t=0}^{C-1}\), form the context set \(\mathcal D_C\), representing the observed history. A contiguous future segment of length \(M\), \(\{(x_t,y_t)\}_{t=C}^{C+M-1}\), is then selected as the prediction target, where \(M\) is the sampled horizon. Following the original NP and ANP formulation~\cite{Garnelo2018,Kim2019}, the full target set is constructed such that \(\mathcal D_C \subseteq \mathcal D_T\).

By randomizing both the context size \(C\) and the target horizon \(M\) across training iterations, FiLM-ANP is exposed to a wide range of conditioning scenarios, from short histories with long horizons to long histories with short horizons. Over many such episodes, the model learns a conditional distribution over load functions that can be queried with varying amounts of observed history. This matches the intended deployment setting, where the available context changes dynamically and forecasts must be produced for different horizons without retraining.

\subsection{Label Supervision and FiLM Conditioning}

The K-means soft labels described in Section~4 are used both as a conditioning signal and as a weak supervision signal for the label inference network. For each load profile \(x\), clustering yields a \(K\)-dimensional soft membership vector
\[
p^{\mathrm{km}}(x) = \big(p^{\mathrm{km}}_1(x), \dots, p^{\mathrm{km}}_K(x)\big),
\]

Constructed from a temperature-scaled softmax over normalized distances. In our experiments, we directly use these probabilities as the teacher distribution,
\begin{equation}
p^{\mathrm{teach}}(x) = p^{\mathrm{km}}(x).
\end{equation}

As shown in Section~\ref{sec:dataset}, the resulting labels exhibit perfect agreement with the corresponding hard assignments while retaining moderate entropy, providing a supervisory signal that is both informative and non-degenerate.

The label inference network produces logits for a context-conditioned prior distribution \(p_\psi(c \mid \mathcal{D}_C)\) and a target-conditioned posterior distribution \(q_\psi(c \mid \mathcal{D}_T)\). These predictions are supervised against \(p^{\mathrm{teach}}(x)\) using a cross-entropy loss \(\mathcal{L}_{\mathrm{sup}}\), with preference given to the posterior when available. An entropy-based regularization term is applied during the early stages of training to discourage premature collapse onto a single behavioural class.

A natural question is why a separate label inference network is required,
rather than directly using the clustering labels for FiLM conditioning.
The reason is that the K-means labels are computed offline using complete
load profiles and therefore depend on future observations that are not
available at inference time. In realistic forecasting scenarios, the model only observes a partial context
$\mathcal{D}_C = \{(x_t, y_t)\}_{t=0}^{C-1}$ and must infer the behavioural class from this incomplete history. The label inference network learns this mapping by approximating the class distribution $p_\psi(c \mid \mathcal{D}_C)$, thereby enabling behaviour-aware conditioning using only information that is observable at prediction time.

This design also avoids information leakage. Because K-means clustering is performed on complete load profiles, the resulting labels depend on future consumption within each frame. Directly using these labels for FiLM conditioning would therefore provide information unavailable at prediction time. In our framework, clustering outputs are used only as external supervision during training, while FiLM conditioning relies on the model’s inferred class distribution computed from the available context. This preserves causal consistency between training and deployment.

In this sense, the label inference network plays a dual role. First, it
acts as a recognition model that maps partial historical observations to
a distribution over behavioural classes, effectively translating static,
offline clustering annotations into a dynamic latent variable that can be
inferred online. Second, it decouples behavioural discovery from
behavioural conditioning: clustering identifies global structure in
complete load profiles, while the label inference network learns how this
structure manifests under incomplete and causal observations.

For FiLM conditioning, we form a blended class distribution
\begin{equation}
p^{\mathrm{cond}}(x)
=
\alpha\, p^{\mathrm{teach}}(x)
+
(1-\alpha)\, q_\psi(c \mid \mathcal{D}_C),
\end{equation}
where the mixing coefficient $\alpha$ is linearly annealed from $1$ to a
minimum value $\alpha_{\min}=0.15$ during an initial warm-up phase (the
first 60 epochs in our experiments). Early in training, FiLM modulation
is therefore driven almost entirely by the clustering-based teacher,
ensuring stable and semantically grounded conditioning. As training
progresses, the model gradually relies more on its own inferred class
prior, enabling a smooth transition from externally guided conditioning
to self-consistent behavioural inference.

Overall, the label path converts unsupervised behavioural discovery into a causal, online-inferable latent variable that can be safely and effectively integrated into the forecasting process through the decoder-conditioning mechanism. The inferred class signal is first mapped to a conditioning embedding and then used by both a layer-wise HyperFiLM branch and a prompt-token branch, enabling Behaviour-conditioned decoding without information leakage at test time.

\section{Experimental Results}

\subsection{Ablation on Behavioural Conditioning}

We run all experiments on the SGSC two-day profiles described in Section~\ref{sec:dataset}. Each profile has 104 half-hourly measurements and is treated as a sample. To prevent information leakage between households, we split the dataset at the user level: households are randomly assigned to \(70\%\)  training, \(15\%\) validation, and \(15\%\) test sets using a fixed random seed. All profiles from the same household are assigned the same split. This user-disjoint protocol evaluates the model on previously unseen households rather than on new profiles from households observed during training.

To isolate the effect of behavioural conditioning, we perform an ablation study comparing the proposed FiLM-conditioned Neural Process variants against a label-agnostic ANP baseline. This ablation progressively introduces behavioural information into the Neural Process framework: first by removing the label path entirely, then by incorporating deterministic (hard) behavioural assignments, and finally by using probabilistic (soft) behavioural representations.

All models share the same Neural Process–based encoder–decoder architecture with attention, as well as identical data splits, context–target sampling strategy, and optimisation settings. They differ only in how behavioural information is incorporated into the model.

The evaluated variants are:

(i) \textbf{ANP (no labels)}: the label inference path and FiLM-based conditioning are disabled, yielding a standard Attentive Neural Process baseline.

(ii) \textbf{FiLM-ANP--Hard}: behavioural conditioning is enabled using one-hot cluster assignments derived from K-means, providing deterministic class information to the decoder.

(iii) \textbf{FiLM-ANP--Soft}: behavioural conditioning is enabled using soft cluster membership vectors, allowing the model to account for uncertainty in behavioural assignment.

All three models use the same NP-based encoder-decoder with attention, and they are trained on the same data split and function-space setup. The ANP baseline turns off the label-inference and FiLM-conditioning pathways. In contrast, FiLM-ANP-Hard and FiLM-ANP-Soft both use this behavioural-conditioning branch, but with hard and soft cluster supervision, respectively. Other than the type of label supervision, the two FiLM-ANP models have the same architecture, loss terms, optimization settings, and training process. This controlled comparison isolates the effect of behavioural FiLM conditioning, and within that mechanism, compares soft and hard label supervision.

\subsection{Evaluation Protocol}

We evaluate all models in a conditional short-term forecasting setting. Each load profile of length \(L=104\) is treated as an individual forecasting task. For each task, the model is provided with a context set consisting of the first \(C\) half-hourly observations and is required to predict the subsequent \(H\) observations, corresponding to a forecast horizon of \(H\) steps. Context lengths and forecast horizons are varied over a grid subject to the constraint \(C + H \le 104\).

To ensure fair comparison across households with different consumption scales, we calculate all evaluation metrics in the normalised space using the same per-user z-score normalisation as in preprocessing. This approach removes scale effects but keeps temporal patterns and relative changes in consumption.  We mainly analyse performance using horizon-wise and context-wise averages, which show how accuracy and uncertainty change with forecast difficulty and the amount of context available. For completeness, we also report overall performance averaged across all evaluated \((C,H)\) pairs.

Model performance is evaluated using the Mean Absolute Error (MAE) and the Continuous Ranked Probability Score (CRPS), which assess point prediction accuracy and probabilistic forecast quality, respectively.

Table~\ref{tab:horizon_mae_crps} reports horizon-wise MAE and CRPS averaged over all evaluated context lengths \(C \in \{48,\dots,95\}\). For each model and forecast horizon \(H \in \{1,\dots,8\}\), results are averaged across all test profiles and all valid context-horizon pairs.

\begin{table}[H]
\centering
\caption{Horizon-wise MAE and CRPS averaged over all context lengths
$C \in \{48,\dots,95\}$.}
\label{tab:horizon_mae_crps}
\begin{tabular}{cccc}
\toprule
Model & Horizon $H$ & MAE & CRPS \\
\midrule
\multirow{8}{*}{ANP}
 & 1 & 0.3918 & 0.3030 \\
 & 2 & 0.4289 & 0.3272 \\
 & 3 & 0.4524 & 0.3424 \\
 & 4 & 0.4684 & 0.3529 \\
 & 5 & 0.4802 & 0.3607 \\
 & 6 & 0.4890 & 0.3666 \\
 & 7 & 0.4957 & 0.3712 \\
 & 8 & 0.5010 & 0.3749 \\
\midrule
\multirow{8}{*}{FiLM-ANP--Hard}
 & 1 & 0.3735 & 0.2850 \\
 & 2 & 0.4045 & 0.3073 \\
 & 3 & 0.4248 & 0.3217 \\
 & 4 & 0.4388 & 0.3317 \\
 & 5 & 0.4493 & 0.3391 \\
 & 6 & 0.4571 & 0.3447 \\
 & 7 & 0.4633 & 0.3492 \\
 & 8 & 0.4683 & 0.3528 \\
\midrule
\multirow{8}{*}{FiLM-ANP--Soft}
 & 1 & 0.3550 & 0.2767 \\
 & 2 & 0.3924 & 0.3017 \\
 & 3 & 0.4165 & 0.3181 \\
 & 4 & 0.4325 & 0.3291 \\
 & 5 & 0.4439 & 0.3372 \\
 & 6 & 0.4523 & 0.3433 \\
 & 7 & 0.4588 & 0.3481 \\
 & 8 & 0.4641 & 0.3520 \\
\bottomrule
\end{tabular}
\end{table}

Table~\ref{tab:horizon_mae_crps} shows that both FiLM-conditioned variants consistently outperform the plain ANP baseline across all forecast horizons. Averaged over \(H=1\)–\(8\), FiLM-ANP--Soft reduces MAE by \(7.9\%\) and CRPS by \(6.9\%\) relative to ANP, while FiLM-ANP--Hard achieves reductions of \(6.1\%\) and \(6.0\%\), respectively. FiLM-ANP--Soft also improves over the hard-label variant, with additional average reductions of \(1.8\%\) in MAE and \(1.0\%\) in CRPS. The soft-label advantage is largest at short horizons: at \(H=1\), FiLM-ANP--Soft reduces MAE and CRPS by \(9.4\%\) and \(8.7\%\) relative to ANP, and by \(5.0\%\) and \(2.9\%\) relative to FiLM-ANP--Hard. Although the gap narrows as the horizon increases, FiLM-ANP--Soft remains superior across all horizons, including \(H=8\). 

From an ablation perspective, the progression from ANP to FiLM-ANP--Hard and then to FiLM-ANP--Soft indicates that behavioural decoder conditioning is beneficial, and that soft behavioural assignments provide additional gains over hard assignments.

\begin{figure}[t]
\centering
\begin{subfigure}{0.48\linewidth}
  \centering
  \includegraphics[width=\linewidth]{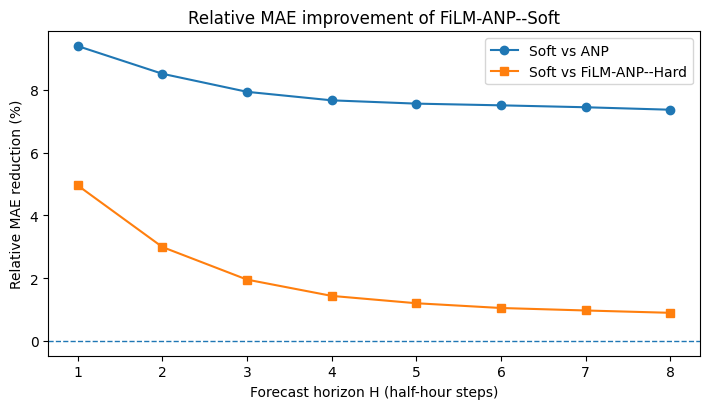}
  \caption{Relative MAE reduction (\%).}
\end{subfigure}
\hfill
\begin{subfigure}{0.48\linewidth}
  \centering
  \includegraphics[width=\linewidth]{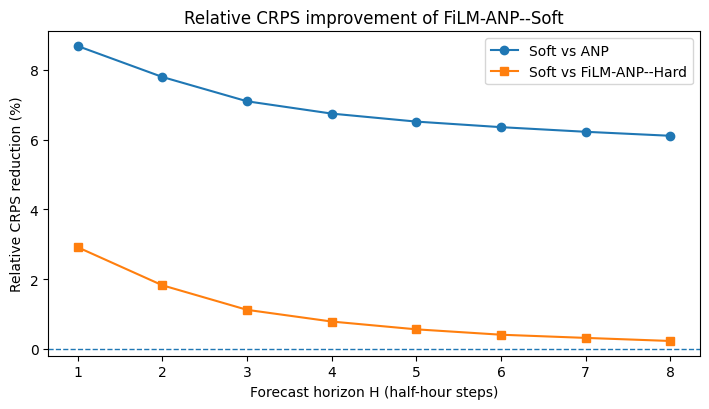}
  \caption{Relative CRPS reduction (\%).}
\end{subfigure}
\caption{Percentage reduction in MAE and CRPS of FiLM-ANP--Soft relative to ANP
and FiLM-ANP--Hard across forecast horizons $H=1$--$8$.}
\label{fig:horizon-rel-improvement}
\end{figure}

Fig.~\ref{fig:horizon-rel-improvement} visualizes the relative reduction in MAE and CRPS achieved by FiLM-ANP--Soft compared with ANP and FiLM-ANP--Hard across forecast horizons. The figure confirms that the largest gains occur at short horizons, while positive improvements are maintained as the horizon increases.

\begin{figure}[t]
    \centering
    \includegraphics[width=0.75\linewidth]{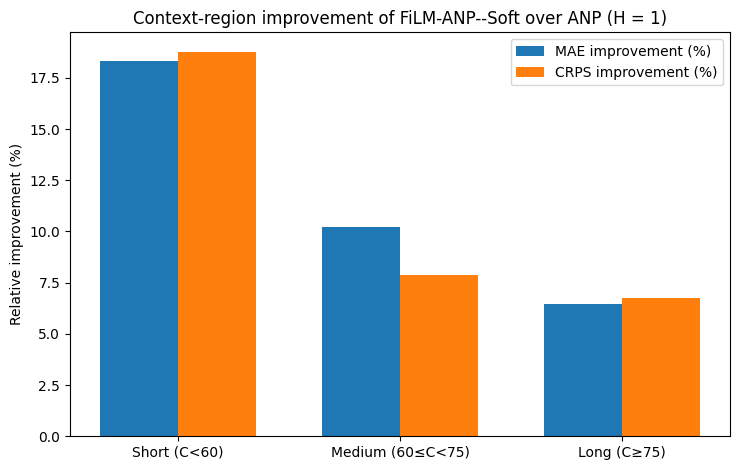}
    \caption{Relative improvement of FiLM-ANP--Soft over ANP as a function of
    context length $C$, grouped into three data-driven regions and evaluated at
    forecast horizon $H=1$.}
    \label{fig:context-region-bar}
\end{figure}

To better understand where soft behavioural labels are most beneficial, we divided context lengths into coarse regions and measured the average relative improvement of FiLM-ANP--Soft over ANP for horizon \(H=1\). Fig.~\ref{fig:context-region-bar} shows that the largest gains occur when historical context is limited: for \(C<60\), FiLM-ANP--Soft reduces MAE by \(18.3\%\) and CRPS by \(18.8\%\). For intermediate contexts (\(60 \le C < 75\)), the gains remain positive but smaller, with reductions of \(10.2\%\) in MAE and \(7.9\%\) in CRPS. For longer contexts (\(C \ge 75\)), the improvements further drop to \(6.4\%\) and \(6.7\%\), respectively. This pattern suggests that soft behavioural labels are most valuable when contextual information is limited.

Across all combinations of context and horizon pairs, the FiLM-conditioned models consistently outperform the plain ANP baseline in both MAE and CRPS. FiLM-ANP--Soft achieves the best overall performance, followed by FiLM-ANP--Hard, indicating that behavioural FiLM conditioning improves both point prediction accuracy and probabilistic forecast quality.  This suggests that behavioural FiLM conditioning improves both the accuracy of point predictions and the quality of probabilistic forecasts.

\subsection{Comparison with Conventional Forecasting Methods and Training Paradigms}

To assess the effectiveness of the proposed NP framework for short-term load forecasting (STLF), we compare it with a set of widely used conventional forecasting methods. The methods cover several common modelling paradigms, including persistence forecasting, regularized linear regression (Elastic Net), kernel-based regression (SVR), ensemble tree methods (XGBoost and LightGBM), feedforward neural networks (MLP), and recurrent neural networks (LSTM), all of which are standard choices in residential STLF studies~\citep{Kong2017,ibrahim2022mlstlf}. For SVR, we use a scalable approximation based on Nyström features with an RBF kernel, followed by an \(\epsilon\)-insensitive linear regressor.This approach provides a computationally efficient alternative to standard kernel SVR for the large supervised training set ~\citep{williams2001}.

To make the comparison fair, all models use the same SGSC household time series, the same preprocessing steps with per-user normalization, and the same user-disjoint splits for training, validation, and testing. Hyperparameters for the baseline models are chosen based on validation performance under the same conditions. For the NP models, the data are grouped into two-day load profiles of length L=104 with a one-day stride, and each profile is treated as a separate forecasting task. During training, the NP models see different context and target setups by randomly splitting each profile into observed context points and prediction targets. As a result, a single profile can contribute multiple conditional prediction instances across training iterations, reflecting the function-learning paradigm of NPs.

For the conventional baselines, we keep the same user-level splits, but represent the data using a standard supervised regression format. Each sample takes a fixed window of 48 half-hourly observations (one day) as input, with the load at forecast horizon \(H\) as the target. This setup matches the minimum NP context length and is common in short-term load forecasting. All models get the same input information, since the minimum NP context size is the same as the fixed input window for the supervised baselines. We generate samples with unit stride, which creates overlapping supervised examples while keeping the temporal order. Using the direct multi-step approach, we train a separate baseline model for each horizon \(H \in \{1,\dots,8\}\). In contrast, the NP-based models work on profile-level tasks and use a single trained model for all horizons by changing the context and target split at inference. Since the conventional baselines produce deterministic predictions, the comparison is based on point forecast metrics (MAE and RMSE).

Both paradigms use the same household time series and data splits, but their training samples are constructed differently. Conventional baselines use supervised input-target pairs created with a sliding window, producing about 4.88 million overlapping samples, each linked to a single forecast target. NP-based models, on the other hand, are trained on about 101,000 load profiles made with a one-day stride. Each profile provides several context-target setups through random partitioning during training. So, conventional models see many overlapping regression samples, while NP models see fewer profile-level samples but make multiple predictions per profile. This difference reflects the distinction between supervised regression and conditional function modelling, rather than any difference in the underlying data.

\begin{table}[t]
\centering
\setlength{\tabcolsep}{3pt}
\renewcommand{\arraystretch}{0.9}
\caption{Horizon-wise MAE / RMSE comparison between FiLM-ANP--Soft and conventional models (H=1--8).}
\label{tab:comparison_conventional_models_compact}

\begin{tabular}{lcccccccc}
\toprule
 & \multicolumn{8}{c}{Horizon $H$} \\
\cmidrule(lr){2-9}
Model 
& 1 & 2 & 3 & 4 & 5 & 6 & 7 & 8 \\
\midrule

FiLM-ANP--Soft (MAE)
& 0.3550 & 0.3924 & 0.4165 & 0.4325 & 0.4439 & 0.4523 & 0.4588 & 0.4641 \\
FiLM-ANP--Soft (RMSE)
& 0.5592 & 0.6145 & 0.6484 & 0.6717   & 0.6894 & 0.7036 & 0.7155 & 0.7258 \\
\midrule

Persistence (MAE)
& 0.3824 & 0.4343 & 0.4757 & 0.5075 & 0.5353 & 0.5587 & 0.5796 & 0.5981 \\
Persistence (RMSE)
& 0.8071 & 0.9090 & 0.9739 & 1.0212 & 1.0592 & 1.0905 & 1.1174 & 1.1407 \\
\midrule

XGBoost (MAE)
& 0.3641 & 0.4025 & 0.4267 & 0.4430 & 0.4549 & 0.4640 & 0.4712 & 0.4769 \\
XGBoost (RMSE)
& 0.6927 & 0.7472 & 0.7782 & 0.7983 & 0.8127 & 0.8233 & 0.8317 & 0.8383 \\
\midrule

LightGBM (MAE)
& 0.3655 & 0.4034 & 0.4273 & 0.4434 & 0.4553 & 0.4642 & 0.4713 & 0.4770 \\
LightGBM (RMSE)
& 0.6943 & 0.7483 & 0.7789 & 0.7988 & 0.8129 & 0.8234 & 0.8316 & 0.8381 \\
\midrule

SVR (MAE)
& 0.3817 & 0.4057 & 0.4214 & 0.4318 & 0.4394 & 0.4450 & 0.4493 & 0.4528 \\
SVR (RMSE)
& 0.7931 & 0.8374 & 0.8624 & 0.8781 & 0.8891 & 0.8969 & 0.9026 & 0.9072 \\
\midrule

Elastic Net (MAE)
& 0.3844 & 0.4247 & 0.4498 & 0.4666 & 0.4788 & 0.4880 & 0.4952 & 0.5009 \\
Elastic Net (RMSE)
& 0.7154 & 0.7703 & 0.8003 & 0.8194 & 0.8330 & 0.8430 & 0.8507 & 0.8568 \\
\midrule

MLP (MAE)
& 0.3654 & 0.4059 & 0.4282 & 0.4431 & 0.4549 & 0.4649 & 0.4760 & 0.4794 \\
MLP (RMSE)
& 0.6930 & 0.7468 & 0.7775 & 0.7980 & 0.8126 & 0.8230 & 0.8313 & 0.8382 \\
\midrule

LSTM (MAE)
& 0.3563 & 0.4014 & 0.4212 & 0.4475 & 0.4554 & 0.4651 & 0.4689 & 0.4843 \\
LSTM (RMSE)
& 0.6852 & 0.7431 & 0.7753 & 0.7967 & 0.8118 & 0.8221 & 0.8317 & 0.8382 \\
\bottomrule
\end{tabular}
\end{table}

Tables~\ref{tab:comparison_conventional_models_compact} summarizes the horizon-wise MAE and RMSE of FiLM-ANP--Soft and the conventional baselines for \(H=1\)–\(8\). Across all evaluated horizons, FiLM-ANP--Soft achieves the lowest RMSE and the lowest MAE at short horizons (\(H=1\)–\(3\)). Averaged across horizons, it also attains the best overall MAE, although the margin over the strongest deterministic baselines is small, with SVR achieving slightly lower MAE at several longer horizons.

A clearer distinction between models is observed in RMSE, where FiLM-ANP--Soft consistently achieves the lowest values across all horizons. Compared to the strongest deterministic baselines, the RMSE reduction ranges from approximately \(13.4\%\) at shorter horizons to \(18.4\%\) at longer horizons. Because RMSE places greater weight on large deviations than MAE, this suggests that the proposed model is particularly effective at reducing larger prediction errors. This is practically relevant in short-term load forecasting, where large errors occur more often during volatile demand periods and can have a disproportionate impact on operational decisions such as reserve allocation, market participation, and demand response~\citep{Hong2016,Morales2014,Bessa2012}.

Overall, these results suggest the benefit of adaptive, context-conditioned forecasting for residential STLF. The proposed NP framework uses a single global model trained across heterogeneous households and evaluated under user-disjoint generalisation. Unlike conventional global regressors with fixed input--output mappings, it models a conditional distribution over load trajectories and conditions predictions on the available context without requiring user-specific retraining.

\subsubsection{Qualitative Analysis of Context-Dependent Forecast Behaviour}
\begin{figure}[H]
    \centering
    \includegraphics[width=\textwidth]{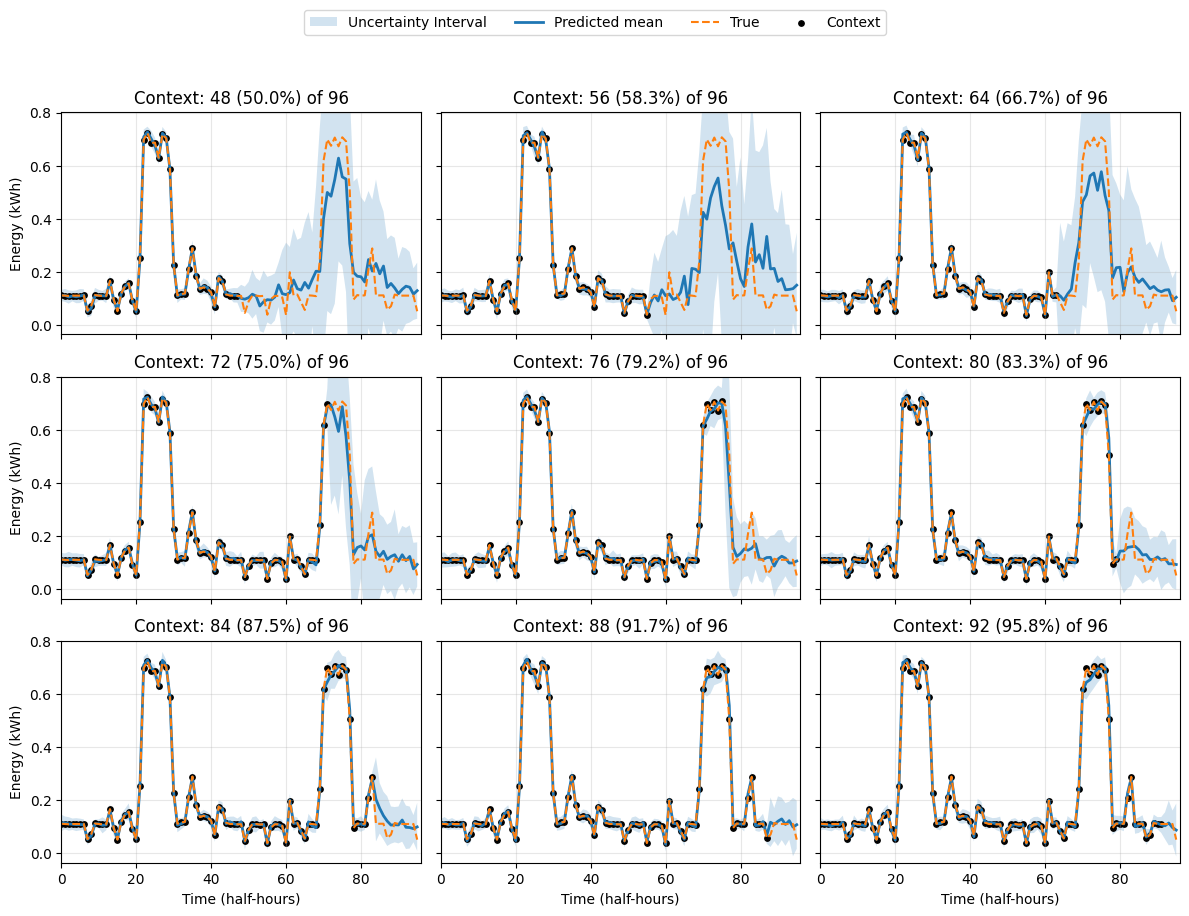} 
    \caption{Context-driven forecasting progression for the FiLM-ANP--Soft model under varying context sizes $C$. As additional observations become available, the predictive mean progressively aligns with the ground truth while predictive uncertainty contracts, illustrating adaptive behaviour under heterogeneous information availability.}
    \label{fig:evolving_context_softanp}
\end{figure}

To further illustrate the adaptive forecasting behaviour beyond aggregated metrics, Fig.~\ref{fig:evolving_context_softanp} presents an example trajectory under progressively increasing context sizes, demonstrating how the model incrementally reconstructs the target curve while maintaining calibrated predictive uncertainty.

\section{Discussion}

The experimental results show that incorporating inferred behavioural structure into the forecasting mechanism improves over the label-agnostic ANP baseline. The behaviour-conditioned variants achieve consistent gains across forecast horizons and context settings, with the strongest performance obtained by the soft-label variant. These results suggest that household-level heterogeneity can provide useful information when it is represented as a context-inferred behavioural signal, rather than treated only as variation to be absorbed by a single global model. The larger gains under limited context availability further indicate that behavioural conditioning is most helpful when the observed history provides only partial evidence about the current load pattern.

The clustering-derived profile information is therefore used as weak supervision, not as ground-truth behavioural labels. These proxy labels do not identify fixed household types or user identities; instead, they describe behavioural structure at the profile level and may vary from one forecasting task to another. Because the K-means assignments are computed from complete load profiles, using them directly at prediction time would leak information from future observations. To avoid this, the label inference network learns to infer a distribution over behavioural classes from the available context. In this way, clustering is used to discover useful structure during training, while test-time decoder conditioning depends only on information observed before the forecast. This preserves Behaviour-conditioned forecasting without relying on complete-profile labels at prediction time.

The comparison between the hard- and soft-label variants suggests that representing uncertainty in behavioural assignment can be beneficial for residential load forecasting. Hard assignments associate each profile with a single behavioural archetype, whereas soft assignments allow a profile to reflect partial membership across multiple regimes. This is useful because household consumption patterns are not always cleanly separable and may contain mixed or transitional behaviours. In the proposed decoder, the inferred behavioural distribution is mapped to a conditioning representation that modulates the predictive pathway, following the general principle of feature-wise conditioning introduced by FiLM~\cite{Perez2018}. The stronger performance of FiLM-ANP--Soft suggests that, in this setting, smoother behaviour-aware conditioning is preferable to imposing a single discrete behavioural mode.

The results also relate to the broader distinction between local, global, and partitioned forecasting strategies. Global models can benefit from statistical sharing across related series, but strong heterogeneity may require mechanisms that preserve series- or regime-specific variation~\cite{montero2021principles}. This view is also consistent with recent heterogeneity-aware time-series modelling, where context- or entity-specific mechanisms are introduced to avoid forcing diverse series into a single undifferentiated representation~\cite{Dong2024HimNet}. The proposed framework follows this principle without training separate local or cluster-specific models. Instead, all profiles share a single probabilistic Neural Process model, while the decoder is conditioned on a context-inferred behavioural representation. This can be interpreted as a soft internal form of heterogeneity-aware adaptation: the model retains shared population-level learning, but allows the predictive distribution to vary according to the behavioural structure inferred for each forecasting task.

Another implication of the results is that the proposed framework supports a single-model evaluation across varying context lengths and forecast horizons. In contrast to fixed-window baselines trained separately for each horizon, the Neural Process formulation exposes the model during training to multiple context--target configurations. The daily-profile construction with a 48-step stride further supports this function-space evaluation by exposing the model to diverse user-day forecasting tasks while limiting excessive overlap between adjacent samples. This allows the same global probabilistic model to generate forecasts under different amounts of observed history and different prediction lengths. The consistent improvements over the label-agnostic ANP baseline across the context--horizon grid suggest that behaviour-aware conditioning complements this function-space training regime, particularly when the available context is limited.

Several limitations should be noted. First, the behavioural labels used in this study are proxy labels derived from K-means clustering, rather than externally validated semantic categories. They provide a practical representation of recurring profile structure, but should not be interpreted as definitive household behaviour types. Second, although the experiments cover variable context lengths and multiple forecast horizons, they do not explicitly evaluate robustness under missing-data or irregular-sampling scenarios. Third, the present evaluation focuses on the overall effect of behaviour-conditioned forecasting within the proposed architecture; future work could examine alternative behavioural representations, richer behavioural annotations, and more detailed analyses of conditioning design choices.

Overall, the results support the central premise of this work: inferred behavioural structure can improve context-conditioned probabilistic residential STLF when it is incorporated inside the forecasting mechanism rather than used only as an external grouping signal. By combining Neural Process function-space learning with weakly supervised behavioural inference and decoder conditioning, the proposed framework provides a unified way to model shared load-function uncertainty and behaviour-specific variation across heterogeneous households.

\section{Conclusion}

This paper investigated whether inferred behavioural structure can be incorporated directly into a Neural Process-based probabilistic forecasting model for residential short-term load forecasting under heterogeneous consumption patterns. We proposed a behaviour-conditioned Attentive Neural Process framework that treats each load profile as a forecasting task conditioned on the available context observations. The model combines a continuous latent variable for shared functional uncertainty with a discrete behavioural latent variable inferred from context and used to condition the decoder.

To support behavioural conditioning when ground-truth behavioural labels are unavailable, clustering-derived profile information is used as weak supervision during training. The resulting proxy labels are not treated as fixed household categories; instead, they provide a practical signal for learning recurring behavioural structure in load profiles. At test time, conditioning relies only on class distributions inferred from the available context, so the model can use behavioural information without directly using complete-profile cluster labels at prediction time.

Experiments on the SGSC residential smart-meter dataset show that the proposed behaviour-conditioned variants consistently improve MAE and CRPS over the label-agnostic ANP baseline across forecast horizons and context settings. The soft-label variant achieves the strongest overall performance, with average reductions of 7.9\% in MAE and 6.9\% in CRPS relative to ANP. Compared with fixed-window deterministic baselines, the same variant achieves lower RMSE across all evaluated horizons while maintaining competitive MAE, suggesting fewer large prediction deviations under heterogeneous consumption patterns.

Overall, the findings demonstrate the value of integrating inferred behavioural structure directly into the forecasting mechanism, rather than treating it only as an external grouping signal. By combining Neural Process-based function-space learning with weakly supervised behavioural inference and decoder conditioning, the proposed framework offers a unified probabilistic approach for capturing shared functional uncertainty and behaviour-specific variation in residential STLF. Future work will consider richer behavioural annotations, evaluation on broader residential datasets, explicit missing-data settings, and more detailed analysis of alternative conditioning strategies.

\section*{CRediT authorship contribution statement}

Ramin Soleimani: Conceptualization, Methodology, Software, Investigation, Validation, Visualization, Writing -- original draft.

Andrea Visentin: Supervision, Methodology, Writing -- review and editing.

Dirk Pesch: Supervision, Conceptualization, Methodology, Project administration, Funding acquisition, Writing -- review and editing.

\section*{Acknowledgements}

This publication has emanated from research supported by grants from Research Ireland under Grant numbers 12/RC/2289-P2 and 18/CRT/6222, which are co-funded under the European Regional Development Fund. For the purpose of Open Access, the author has applied a CC BY public copyright licence to any Author Accepted Manuscript version arising from this submission.

\section*{Declaration of competing interest}

The authors declare that they have no known competing financial interests or personal relationships that could have appeared to influence the work reported in this paper.

\section*{Data availability}

The Smart Grid, Smart City (SGSC) customer trial data used in this study are publicly available through the Australian Government open-data portal. Processed experimental outputs and code may be made available by the corresponding author upon reasonable request.

\section*{Declaration of generative AI and AI-assisted technologies in the manuscript preparation process}

During the preparation of this work, the authors used ChatGPT to improve the readability and language clarity of the manuscript. After using this tool, the authors reviewed and edited the content as needed and take full responsibility for the content of the article.

\end{document}